\documentclass[letterpaper]{article} % DO NOT CHANGE THIS
\usepackage{aaai2026}  % DO NOT CHANGE THIS
\usepackage{times}  % DO NOT CHANGE THIS
\usepackage{helvet}  % DO NOT CHANGE THIS
\usepackage{courier}  % DO NOT CHANGE THIS
\usepackage[hyphens]{url}  % DO NOT CHANGE THIS
\usepackage{graphicx} % DO NOT CHANGE THIS
\usepackage{natbib}  % DO NOT CHANGE THIS AND DO NOT ADD ANY OPTIONS TO IT
\usepackage{caption} % DO NOT CHANGE THIS AND DO NOT ADD ANY OPTIONS TO IT
\usepackage{algorithm}
\usepackage{algorithmic}
\usepackage{newfloat}
\usepackage{listings}
\DeclareCaptionStyle{ruled}{labelfont=normalfont,labelsep=colon,strut=off} % DO NOT CHANGE THIS
\floatstyle{ruled}
\newfloat{listing}{tb}{lst}{}
\floatname{listing}{Listing}
\usepackage{colortbl}
\usepackage[table, xcdraw, dvipsnames]{xcolor}
\usepackage{mdframed}

\usepackage{bm}
\usepackage{booktabs}

\usepackage{pifont} % 提供特殊符号
\usepackage{tcolorbox}
\definecolor{problemblue}{rgb}{0.6157, 0.7647, 0.9020}
\definecolor{problemyellow}{rgb}{1, 0.8510, 0.4}
\usepackage{amsmath,amssymb,amsfonts}
\usepackage{enumitem}
\usepackage{cleveref}
\newcommand{\fedavg}{{FedAvg}}
\newcolumntype{I}{!{\vrule width 1pt}}
\definecolor{lightgray}{gray}{.9}
\definecolor{deepgray}{gray}{.8}
\definecolor{mygray}{gray}{.9}

\definecolor{problemblue}{rgb}{0.6157, 0.7647, 0.9020}
\definecolor{problemyellow}{rgb}{1, 0.8510, 0.4}

\newcommand{\pub}[1]{{\color{gray}{\small{[{#1}]}}}}
\usepackage{multirow}
\makeatletter
\newcommand{\thickhline}{%
    \noalign {\ifnum 0=`}\fi \hrule height 1pt
    \futurelet \reserved@a \@xhline
}
\makeatother
\newcommand{\fedprox}{{FedProx}}

\newcommand{\moon}{{MOON}}
\newcommand{\FedPLVM}{{FedPLVM}}

\newcommand{\fedproto}{{FedProto}} 
\newcommand{\harmof}{{HarmoFL}}

\newcommand{\FPL}{{FPL}}

\newcommand{\feduv}{{FedUV}}

\newcommand{\cfdlplabbv}{{CDPA}}
\newcommand{\lfbsnabbv}{{FSR}}

\usepackage{pifont}
\usepackage{fontawesome}

\title{Divide, Conquer and Unite: Hierarchical Style-Recalibrated Prototype Alignment for Federated Medical Segmentation}
\title{Divide, Conquer and Unite: Hierarchical Style-Recalibrated Prototype Alignment for Federated Medical Segmentation}
\author {
    Xingyue Zhao\textsuperscript{\rm 1}\equalcontrib,
    Wenke Huang\textsuperscript{\rm 1}\equalcontrib,
    Xingguang Wang\textsuperscript{\rm 2}\equalcontrib,
    Haoyu Zhao\textsuperscript{\rm 1},
    Linghao Zhuang\textsuperscript{\rm 3},
    Anwen Jiang\textsuperscript{\rm 4},
    Guancheng Wan\textsuperscript{\rm 1},
    Mang Ye\textsuperscript{\rm 1}\thanks{Corresponding author: yemang@whu.edu.cn}
}
\affiliations {
    \textsuperscript{\rm 1} School of Computer Science, Wuhan University, China\\
    \textsuperscript{\rm 2}School of Software Engineering, Xi'an Jiaotong University, China\\
    \textsuperscript{\rm 3}School of Software Engineering, Xinjiang University, China\\
    \textsuperscript{\rm 4}College of Mathematics and System Science, Xinjiang University, China
}
\usepackage{bibentry}
\begin{document}

\maketitle

\begin{abstract}
Federated learning enables multiple medical institutions to train a global model without sharing data, yet feature heterogeneity from diverse scanners or protocols remains a major challenge. Many existing works attempt to address this issue by leveraging model representations (e.g., mean feature vectors) to correct local training; however, they often face two key limitations: 1) Incomplete Contextual Representation Learning: Current approaches primarily focus on final-layer features, overlooking critical multi-level cues and thus diluting essential context for accurate segmentation. 2) Layerwise Style Bias Accumulation: Although utilizing representations can partially align global features, these methods neglect domain-specific biases within intermediate layers, allowing style discrepancies to build up and reduce model robustness. To address these challenges, we propose FedBCS to bridge feature representation gaps via domain-invariant contextual prototypes alignment. Specifically, we introduce a frequency-domain adaptive style recalibration into prototype construction that not only decouples content-style representations but also learns optimal style parameters, enabling more robust domain-invariant prototypes. Furthermore, we design a context-aware dual-level prototype alignment method that extracts domain-invariant prototypes from different layers of both encoder and decoder and fuses them with contextual information for finer-grained representation alignment. Extensive experiments on two public datasets demonstrate that our method exhibits remarkable performance.
\end{abstract}
\begin{links}
    \link{Code}{https://github.com/zxy1234321/FedBCS}
\end{links}

\begin{figure}[t]
\centering
\includegraphics[width=\linewidth]{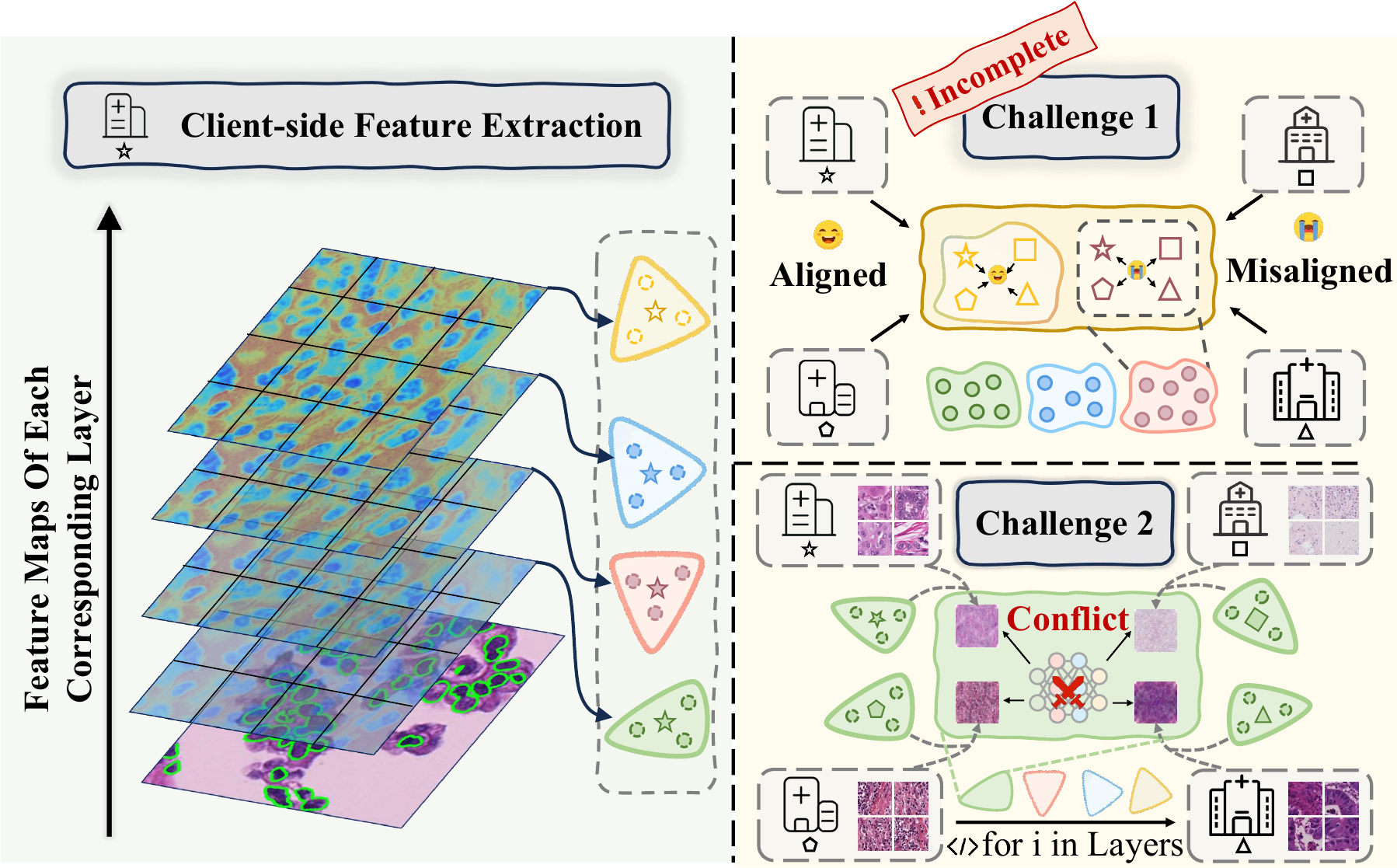}
% \vspace{-10pt}
\captionsetup{font=small}
\caption{\small{
\textbf{Problem illustration} of existing methods under domain skew setting. The top row highlights two major challenges—\textbf{incomplete contextual representation learning} and \textbf{layerwise style bias accumulation}. 
}}
\label{fig:problem}
% \vspace{-15pt}
\end{figure}

\section{Introduction}
Medical image segmentation is crucial for computer-assisted diagnosis, and most deep learning-based techniques require large amounts of data for effective training~\cite{wang2024cardiovascular,zhao2024ultrasound, zhao2024sam}. However, sharing such data among hospitals is often restricted by privacy concerns and regulatory constraints. Federated learning has emerged as a promising strategy for multi-site collaborative training of a global model without compromising patient privacy. A foundational method in this domain, FedAvg~\cite{FedAvg_AISTATS17}, aggregates model parameters from different clients to form a global model, which is then redistributed to each client for further training. Despite these advances, data heterogeneity remains a significant challenge in real-world scenarios~\cite{FLSurveyandBenchmarkforGenRobFair_TPAMI24, zhao2024morestyle}. Datasets from different medical sites are often non-IID due to variations in scanners and imaging protocols, leading to different feature distributions across clients~\cite{jiang2022harmofl, wang2025pfedsam, zhao2024wia, zhuang2025activessf}. This diversity can hinder model convergence, reduce generalization capabilities, and result in suboptimal performance.

To address data heterogeneity, researchers use model representations to correct local training and improve generalization, employing mean feature vectors for privacy~\cite{MOON_CVPR21,FedProx_MLSys2020}. Subsequent work then introduces prototypes (class-specific mean feature vectors) and federated prototype learning, where client prototypes minimize distances to global or same-class prototypes while maximizing separation from different-class ones~\cite{FedProto_AAAI22,FPL_CVPR23}.
Many researchers have also introduced specialized strategies to further improve training in these settings. For example, one method clusters client features to generate multiple prototypes per domain instead of a single average, and refines feature-level distance by elevating it to an $\alpha$ power, addressing unequal learning due to domain diversity~\cite{FedPLVM_NeurIPS24}. 
Another study leverages Gaussian Mixture Models to generate prototypes and soft predictions, then weights these outputs based on both quality and quantity~\cite{FedGMKD_NeurIPS24}. 

In federated learning across multiple medical institutions, variations in scanning protocols, staining methods, and patient populations often amplify feature distribution differences~\cite{jiang2023fair,chen2024think, li2025expert}. Despite recent advances, many existing approaches still depend on a single-layer or simplified representation, making it challenging to fully capture the multi-scale contextual complexities and style variations in medical image segmentation, where distribution shifts can be more pronounced. We group these limitations into two key challenges and highlight them in Figure~\ref{fig:problem}:
\definecolor{cadetblue}{RGB}{95,158,160} % Define CadetBlue color
\definecolor{keywordcolor}{RGB}{178,34,34} % Define FireBrick color for keywords

\begin{mdframed}[backgroundcolor=cadetblue!10, linewidth=0.8pt, linecolor=cadetblue!80, roundcorner=5pt]
\textbf{\textcolor{keywordcolor}{Layerwise Style Bias Accumulation}}:

\textbf{Style discrepancies from different medical protocols accumulate through intermediate layers, degrading cross-domain feature alignment.} Many existing methods only normalize style at input level or align final-layer features~\cite{FPL_CVPR23}, leaving intermediate-layer style biases unaddressed and hindering model generalization.
\end{mdframed}

\begin{mdframed}[backgroundcolor=cadetblue!10, linewidth=0.8pt, linecolor=cadetblue!80, roundcorner=5pt]
\textbf{\textcolor{keywordcolor}{Incomplete Contextual Representation Learning}}:

\textbf{Single-layer feature alignment fails to capture the complete semantic hierarchy needed for medical image segmentation.} While lower layers encode crucial local details and higher layers capture abstract structures~\cite{li2023d}, relying solely on final-layer representations leads to incomplete semantic understanding and compromised performance~\cite{sung2024contextrast}.
\end{mdframed}

Considering these issues, we propose a novel method called FedBCS, specifically designed for feature heterogeneous federated medical image segmentation.  
For \textbf{Layerwise Style Bias Accumulation} challenge, we propose a Frequency-domain Style Recalibration (FSR) module, specifically designed for constructing domain-invariant prototypes in federated learning. We explore methods for decoupling and recalibrating domain-specific style variations during federation by learning adaptive style parameters from cross-client feature representations. Given that style variations propagate through network layers and affect prototype quality differently, the FSR module modulates feature styles in the frequency domain for robust prototype construction. First, we decompose intermediate features into content and style components via Fourier Transform, where phase spectra preserve semantic content while amplitude spectra encode style information~\cite{lee2023decompose,chen2021amplitude}. Building on this, we introduce learnable style parameters to adaptively recalibrate the normalized amplitude during federation, enabling optimal style modulation while maintaining semantic consistency. This strategy effectively eliminates domain-specific style biases while ensuring robust prototype construction across different domains.

\textbf{Once robust domain-invariant prototypes are constructed, effectively leveraging multi-level semantic information becomes crucial for accurate feature alignment.} Building upon these domain-invariant prototypes, we further address the challenge of \textbf{Incomplete Contextual Representation Learning}. Current methods typically construct prototypes from single-layer features, leading to incomplete semantic representations. To tackle this limitation, we propose a context-aware dual-level prototype alignment (CDPA) strategy that leverages both encoder and decoder features. Specifically, we extract domain-invariant prototypes from multiple network layers and design a hierarchical fusion mechanism to integrate multi-scale contextual information. This approach enables more comprehensive semantic alignment by capturing both low-level details and high-level semantic patterns, ultimately achieving finer-grained feature representation. Our main contributions are summarized below.
% 在正文中使用
\begin{itemize}

\item[{\ding{182}}] \textbf{Revealing Federated Learning Limitations in Heterogeneous FL.} We identify two critical limitations in Heterogeneous federated medical segmentation: layerwise style bias accumulation and incomplete contextual representation learning, which significantly impact cross-client feature alignment.

\item[{\ding{183}}] \textbf{Novel Federated Prototype Learning Framework.} We propose FedBCS that integrates frequency-domain style recalibration for robust prototype construction with context-aware dual-level alignment to achieve comprehensive semantic representation learning.
\item[{\ding{184}}] \textbf{Theoretical and Empirical Validation.} We provide theoretical convergence guarantees and demonstrate superior empirical performance through extensive experiments on two medical segmentation tasks.

\end{itemize}

\section{Related Work}
\subsection{Data Heterogeneous Federated Learning}
Federated learning (FL) provides a privacy-preserving framework for collaborative model training without data centralization~\cite{konevcny2016federated}. The foundational method FedAvg~\cite{FedAvg_AISTATS17} iteratively builds a global model by aggregating locally trained parameters. While effective in homogeneous settings, FedAvg often struggles with data heterogeneity, which frequently arises in medical contexts where varying imaging protocols, staining methods, or patient populations lead to significant divergence across institutions~\cite{wang2022personalizing}. Based on FedAvg, numerous methods have attempted to address the data heterogeneity problem from different perspectives~\cite{FLSurveyandBenchmarkforGenRobFair_TPAMI24}. Some approaches focus on training a single global model, aiming to refine the global parameters for improved cross-domain performance~\cite{FPL_CVPR23,huang2022learn, GGEUR_CVPR25}, while others emphasize fairness, ensuring balanced outcomes across heterogeneous clients. Among these, individual/group fairness~\cite{ezzeldin2023fairfed}, collaboration fairness~\cite{jiang2023fair}, and performance fairness~\cite{chen2024fair,li2021ditto} are three widely studied notions. Meanwhile, personalization methods~\cite{FedBN_ICLR21,huang2023generalizable} learn individualized models tailored to each client’s domain, and domain generalization approaches~\cite{FedDG_CVPR21,zhang2023federated} seek robust performance on unseen distributions. In this paper, we address the challenge of data heterogeneity across different medical institutions and concentrate on training a single global model that can robustly perform across these diverse domains.
\subsection{Model Representation Alignment in Federated Learning}
In Federated Learning (FL), many methods rely on model representations to enhance local training in heterogeneous data settings~\cite{wan2024federated,liao2024rethinking, ye2023heterogeneous}. However, sharing raw features can pose privacy concerns. To address this, mean feature vectors are used to retain essential information while minimizing data leakage. Mean feature vectors computed from same-class samples are called prototypes. MOON~\cite{MOON_CVPR21} and FedProx~\cite{FedProx_MLSys2020} utilize contrastive learning to maximize the agreement between local and global model mean feature vectors, thereby aligning local representations with the global model. FedProto~\cite{FedProto_AAAI22} enhances the alignment of client-side training by minimizing the distance between client-side class prototypes and global class prototypes. However, by emphasizing single-domain performance under label skew, these approaches often neglect domain shift, thereby limiting their generalization across diverse domains. Thus, many methods adopt prototype-based approaches to address domain skew. For instance, FPL~\cite{FPL_CVPR23} employs server-side, cluster-based prototypes combined with contrastive learning to align same-class representations. However, it relies on a single representation for each client, leading to limited representational capacity and causing unequal learning in heterogeneous domains. By contrast, FedContrast-GPA~\cite{FedContrast-GPA_MICCAI23} and FedPLVM~\cite{FedPLVM_NeurIPS24} cluster features on the client side to produce multiple prototypes per class, mitigating the drawbacks of single-representation strategies. FedGMKD~\cite{FedGMKD_NeurIPS24}, on the other hand, generates prototypes using Gaussian Mixture Models. Despite their effectiveness, these methods rely solely on final-layer encoder features and overlook the influence of domain-specific style differences on feature representations. In this work, we focus on finer-grained feature alignment and explicitly handling style-induced variations to enhance federated segmentation across diverse medical domains.
\begin{figure*}[t]
% 	\vspace{-4pt}
	\begin{center}
    \includegraphics[width=0.95\linewidth]{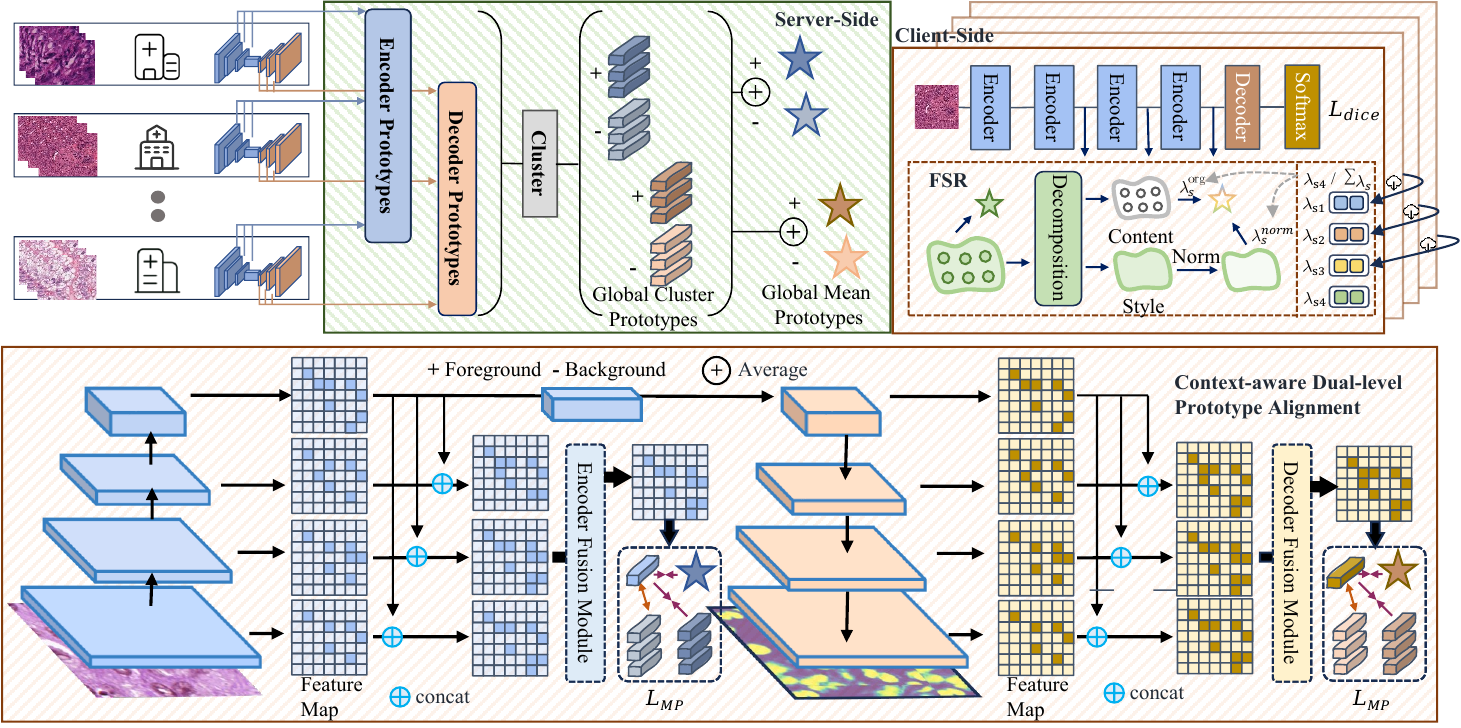}
		% \put(-109,93){\scriptsize\cref{eq:celoss}}
		% \put(-124,55){\scriptsize\cref{eq:cpclloss}}
		% \put(-68,25){\scriptsize\cref{eq:upcrloss}}
	\end{center}
	% \vspace{-5pt}
	\captionsetup{font=small}
    \caption{\small\textbf{Architecture illustration} of FedBCS. To address layerwise style bias accumulation, we implement frequency-domain style recalibration during local prototype construction. To capture complete contextual representations, we extract and align multi-level prototypes from both encoder and decoder pathways. These prototypes are aggregated at the server to derive global prototypes for guiding local training. Zoom in for details.} 
	\label{fig:framework}
	% \vspace{-15pt}
\end{figure*}

\section{Methodology}
\subsection{Preliminaries}
\label{sec:Preliminaries}
Following typical Federated Learning setup~\cite{FedAvg_AISTATS17,FedProx_MLSys2020}, we assume there are \(M\) clients. Each client \(m\) holds a private dataset \(\{(x_i, y_i)\}_{i=1}^{N_m}\), where \(N_m\) denotes the size of client \(m\)'s local dataset. The objective of federated learning is to minimize the weighted average of local objectives:
\begin{equation}\small
\setlength\abovedisplayskip{0pt} \setlength\belowdisplayskip{0pt}
    \begin{split}
    \min_\omega \; F(\omega) = \sum_{m=1}^{M} w_m \, \left( \frac{1}{N_m} \sum_{i=1}^{N_m} L\bigl(x_i, y_i; \omega\bigr) \right),\\
    \end{split}
\end{equation}
where \(w_m\) is the weight of client \(m\), and \(L(x_i, y_i; \omega)\) is the loss for data point \((x_i, y_i)\) under model parameters \(\omega\). Each client updates its model locally and sends the updates to a central server for aggregation.

\noindent\textbf{Data Heterogeneity.} In federated medical image segmentation, while the label distribution \(P(y)\) remains consistent across clients, the feature distribution \(P(x \mid y)\) varies due to different scanning devices and protocols, leading to significant feature heterogeneity across medical institutions.

We adopt a standard segmentation network architecture shared across all clients, consisting of an encoder \(F_{\mathrm{enc}}: X \to Z\) that maps an input image \(x\) to a \(d\)-dimensional feature map \(\mathbf{z}_{\mathrm{enc}} = F_{\mathrm{enc}}(x) \in \mathbb{R}^d\), and a decoder \(F_{\mathrm{dec}}: Z \to O\) that produces a segmentation mask \(\mathbf{o}_{\mathrm{dec}} = F_{\mathrm{dec}}(\mathbf{z}_{\mathrm{enc}}) \in \mathbb{R}^c\), where \(c\) denotes the number of classes.

\subsection{Frequency-domain Style Recalibration for
Prototype Construction}

\textbf{Federated prototype learning.}
Recent federated learning approaches leverage class prototypes alongside network weights for cross-client knowledge sharing. A class prototype represents the mean feature representation of that class on each client:
\begin{equation}\small
\setlength\abovedisplayskip{0pt} \setlength\belowdisplayskip{0pt}
    \begin{split}
p_m^{c}\;=\;\frac{1}{|N_m^{c}|}\sum_{x_i\in N_m^{c}}
F_{\text{enc}}^{\,m}(x_i),
    \end{split}
\end{equation}
where $F_{\text{enc}}^{\,m}(\cdot)$ represents client $m$'s local encoder (architecturally identical across clients but with parameters synchronized only during FedAvg rounds), and $N_m^{c}$ denotes the set of pixels/patches labeled as class $c$ on that client. The server aggregates these local prototypes to guide subsequent local training phases, enabling efficient cross-client semantic alignment while maintaining privacy and minimizing communication costs.

\noindent\textbf{Motivation.} Existing prototype-based methods assume features from the same class follow similar distributions across clients, suggesting simple averaging should suffice for prototype construction. However, in medical imaging, diverse scanning protocols introduce significant style variations that propagate through network layers, manifesting in the amplitude spectra of feature representations. When constructing prototypes through direct feature averaging, this leads to:

\begin{itemize}[leftmargin=1.5em, itemsep=-0.2em, topsep=0em]
\item[\faThumbsODown] \textit{Layerwise Style Bias Accumulation:} Style discrepancies compound through intermediate layers, causing features of the same anatomical structure to appear increasingly different across institutions and undermining the effectiveness of prototype learning.
\end{itemize}

\noindent\raisebox{-0.2em}{\includegraphics[scale=0.25]{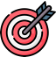}} To establish reliable semantic anchors for cross-client feature alignment, prototypes must be constructed from style-invariant features. This requires effectively removing style variations while preserving semantic information before prototype computation. A promising direction is to leverage frequency domain analysis, where content and style information naturally separate into phase and amplitude components, enabling effective style-invariant prototype construction.

\noindent\textbf{Style-invariant Prototype Generation.} To address the aforementioned challenges in prototype-based federated learning for medical image segmentation, we propose Frequency-domain Style Recalibration that removes domain-specific style variations while preserving semantic information during prototype construction. Our key insight is that \textit{style and content information naturally separate in the frequency domain, allowing for targeted style modulation before prototype generation.} Given the encoder feature map $\mathbf{z}_{\mathrm{enc}} \in \mathbb{R}^{C \times H \times W}$, conventional prototype-based methods construct class-specific prototypes through direct averaging:
\begin{equation}\small
\setlength\abovedisplayskip{0pt} \setlength\belowdisplayskip{0pt}
    \begin{split}
    p_m^c = \frac{1}{|N_m^c|} \sum_{x_i \in N_m^c} F_{\mathrm{enc}}(x_i),
    \end{split}
\end{equation}
% where $N_m^c$ denotes samples of class $c$ in client $m$'s local dataset. 
where $N_m^c$ denotes samples of class $c$ in client $m$'s local dataset, and $F_{\mathrm{enc}}(\cdot)$ represents the encoder network that extracts feature representations. However, these prototypes directly inherit domain-specific style variations from different medical institutions, compromising their effectiveness as semantic anchors for cross-client feature alignment. To address this limitation, we introduce FSR that processes features in the frequency domain before prototype construction. Specifically, given a feature map $\mathbf{z}_{\mathrm{enc}}$, we first decompose it into amplitude and phase components via 2D Fourier transform:
\begin{equation}\small
\setlength\abovedisplayskip{0pt} \setlength\belowdisplayskip{0pt}
\begin{split}
Z(\alpha, \beta) = \frac{1}{HW} \sum_{h=0}^{H-1} \sum_{w=0}^{W-1} z(h, w) \exp\left(-i2\pi \left(\frac{\alpha h}{H} + \frac{\beta w}{W}\right)\right),
\end{split}
\end{equation}
where the amplitude spectrum $\chi$ and phase spectrum $\gamma$ are computed as:
\begin{equation}\small
\setlength\abovedisplayskip{0pt} \setlength\belowdisplayskip{0pt}
\begin{split}
\chi(\alpha, \beta) &= \sqrt{\mathcal{R}(\alpha, \beta)^2 + \mathcal{I}(\alpha, \beta)^2}, \\
\gamma(\alpha, \beta) &= \arctan\left(\frac{\mathcal{I}(\alpha, \beta)}{\mathcal{R}(\alpha, \beta)}\right).
\end{split}
\end{equation}

The amplitude spectrum encodes style information while the phase spectrum preserves semantic content. To achieve adaptive style control, we introduce learnable parameters that balance normalized and original style information:
\begin{equation}\small
\setlength\abovedisplayskip{0pt} \setlength\belowdisplayskip{0pt}
\begin{split}
\lambda_s^{\text{norm}}, \lambda_s^{\text{org}} = \sigma(W_s[\text{GAP}(\chi_{\text{norm}}); \text{GAP}(\chi)]),
\end{split}
\end{equation}
where $\text{GAP}$ denotes global average pooling, $W_s$ are learnable weights, and $\chi_{\text{norm}}$ is the instance-normalized amplitude. The style-recalibrated features are then obtained through inverse Fourier transform:
\begin{equation}\small
\setlength\abovedisplayskip{0pt} \setlength\belowdisplayskip{0pt}
\begin{split}
\hat{z}_{\mathrm{enc}} = \text{IFT}\left(\lambda_s^{\text{norm}} \chi_{\text{norm}} + \lambda_s^{\text{org}} \chi, \gamma \right).
\end{split}
\end{equation}

By applying FSR to the encoder features before averaging, we construct style-invariant prototypes that better preserve class-specific semantic information:
\begin{equation}\small
\setlength\abovedisplayskip{0pt} \setlength\belowdisplayskip{0pt}
\begin{split}
\hat{p}_m^c = \frac{1}{|N_m^c|} \sum_{x_i \in N_m^c} \hat{z}_{\mathrm{enc}},
\end{split}
\end{equation}
where $\hat{z}_{\mathrm{enc}}$ is the style-recalibrated feature obtained through our FSR module. Compared to conventional prototypes that directly inherit domain-specific styles, our FSR-based prototypes better capture the underlying semantic representations while being more robust to cross-domain variations.

\subsection{Context-aware Dual-Level Prototype Alignment}
\label{sec:cfdlpl}
\noindent\textbf{Motivation.}
While our FSR module effectively addresses style variations, existing prototype-based methods~\cite{FedProto_AAAI22,FPL_CVPR23} typically only utilize final-layer features for prototype construction and alignment. This single-layer approach overlooks the rich hierarchical information distributed across network layers, where both local details and global structures are crucial for medical image segmentation. When constructing prototypes for cross-client alignment, this leads to:

\begin{itemize}[leftmargin=1.5em, itemsep=-0.2em, topsep=0em]
\item[\faThumbsODown] \textit{Incomplete Contextual Representation Learning:} Single-layer feature alignment fails to capture the complete semantic hierarchy needed for medical image segmentation, where both fine-grained tissue details and global anatomical structures must be precisely delineated.
\end{itemize}

\noindent\raisebox{-0.2em}{\includegraphics[scale=0.25]{LaTeX/potential.png}} To address this challenge, we explore how to effectively integrate and align prototypes across different semantic levels. This motivates our development of a context-aware prototype alignment strategy that specifically addresses the multi-scale nature of medical image segmentation.

\noindent\textbf{Multi-level Feature Integration.} 
Building upon our FSR module, which generates style-invariant features \(\hat{z}_{\mathrm{enc}}\) through frequency-domain recalibration, we now extend prototype construction to multiple network layers. Given the encoder-decoder architecture with \(K\) layers, for layer \(k \in \{1, \dots, K\}\), we compute class-specific prototypes from the style-recalibrated features:
\begin{equation}\small
\setlength\abovedisplayskip{0pt} \setlength\belowdisplayskip{0pt}
    \begin{split}
    p_{\mathrm{enc}}^{k,c} = \frac{\sum_{v \in \hat{z}_{\mathrm{enc}}^k} v \, \mathbb{I}[g(v) = c]}{|N^c|}, 
    p_{\mathrm{dec}}^{k,c} = \frac{\sum_{v \in \hat{z}_{\mathrm{dec}}^k} v \, \mathbb{I}[g(v) = c]}{|N^c|},
    \end{split}
\end{equation}
where \(g(v)\) maps the feature vector to its ground-truth class label, and \(\mathbb{I}[\cdot]\) is the indicator function. The resulting layer-wise prototypes are denoted as \(\displaystyle P_{\mathrm{enc}}^k = \{ p_{\mathrm{enc}}^{k,1}, \dots, p_{\mathrm{enc}}^{k,c} \}\) and \(\displaystyle P_{\mathrm{dec}}^k = \{ p_{\mathrm{dec}}^{k,1}, \dots, p_{\mathrm{dec}}^{k,c} \}\).

In medical image segmentation, different network layers capture complementary anatomical information. Shallow layers encode local tissue textures and edge patterns crucial for precise boundary delineation, while deeper layers represent high-level anatomical structures and spatial relationships between organs. To effectively combine these multi-scale features, we design a hierarchical fusion strategy. Specifically, we concatenate shallow-layer prototypes with deeper ones:
\begin{equation}\small
\setlength\abovedisplayskip{0pt} \setlength\belowdisplayskip{0pt}
    \begin{split}
    \hat{P}_{\mathrm{enc}}^{k} = \text{Concat}(P_{\mathrm{enc}}^{k}, P_{\mathrm{enc}}^{k+1}),
    \end{split}
\end{equation}
This design allows high-level semantic guidance to enhance local feature discrimination - for instance, knowing the overall organ shape (deep features) helps better determine tissue boundaries (shallow features) in ambiguous regions. 

To mitigate communication overhead while preserving multi-scale information, we introduce a lightweight fusion module \(F: \hat{P} \to Q\) that adaptively integrates features across layers:
\begin{equation}\small
\setlength\abovedisplayskip{0pt} \setlength\belowdisplayskip{0pt}
    \begin{split}
    Q_{\mathrm{enc}} = F(\hat{P}_{\mathrm{enc}}), Q_{\mathrm{dec}} = F(\hat{P}_{\mathrm{dec}}),
    \end{split}
\end{equation}
The fusion module learns to emphasize the most discriminative features at each scale through 1×1 convolutions, resulting in compact yet comprehensive prototypes for each client \(m\):
\begin{equation}\small
\setlength\abovedisplayskip{0pt} \setlength\belowdisplayskip{0pt}
    \begin{split}
    Q_{\mathrm{enc}}^m = \{ q_{\mathrm{enc}}^{m,1}, \dots, q_{\mathrm{enc}}^{m,c} \}, Q_{\mathrm{dec}}^m = \{ q_{\mathrm{dec}}^{m,1}, \dots, q_{\mathrm{dec}}^{m,c} \}.
    \end{split}
\end{equation}

\noindent\textbf{Server-side Prototype Alignment.} 
In federated medical image segmentation, different institutions may focus on different anatomical regions or pathological conditions, leading to varied feature distributions. To achieve consistent cross-client alignment while preserving privacy, we introduce a clustering-based alignment strategy. Using FINCH~\cite{sarfraz2019efficient}, we group similar prototypes from different clients:
\begin{equation}\small
\setlength\abovedisplayskip{0pt} \setlength\belowdisplayskip{0pt}
    \begin{split}
    \text{Cluster}\left( \{ q^{m,c}_{\mathrm{enc}} \}_{m=1}^{M} \right) \longrightarrow \{ q^{m,c}_{\mathrm{enc}} \}_{m=1}^{S} \in \mathbb{R}^{S \times d}\\
    \text{Cluster}\left( \{ q^{m,c}_{\mathrm{dec}} \}_{m=1}^{M} \right) \longrightarrow \{ q^{m,c}_{\mathrm{dec}} \}_{m=1}^{S} \in \mathbb{R}^{S \times d}.
    \end{split}
\end{equation}

To stabilize optimization, we further compute mean prototypes for each class:
\begin{equation}\small
\setlength\abovedisplayskip{0pt} \setlength\belowdisplayskip{0pt}
    \begin{split}
    \bar{q}_{\mathrm{enc}}^c = \frac{1}{S} \sum_{q^c \in Q_{\mathrm{enc}}^c} q^c, \bar{q}_{\mathrm{dec}}^c = \frac{1}{S} \sum_{q^c \in Q_{\mathrm{dec}}^c} q^c,
    \end{split}
\end{equation}
where \(S\) denotes the number of clusters.

\noindent\textbf{Learning Objectives.} 
For effective feature alignment, we design a dual-stream learning objective. Given a medical image sample \((x_i, y_i)\), we extract feature embeddings \(\mathbf{e}^i = \{\mathbf{e}_{\mathrm{enc}}^i, \mathbf{e}_{\mathrm{dec}}^i\}\). The contrastive term encourages alignment with same-class prototypes while maintaining separation from different anatomical regions:
\begin{equation}\small
\setlength\abovedisplayskip{0pt} \setlength\belowdisplayskip{0pt}
    \begin{split}
    L_{contra} = -\log \frac{\sum_{q \in Q^c} e^{\mathrm{Sim}(\mathbf{e}^i, q)}}{\sum_{q \in Q^c} e^{\mathrm{Sim}(\mathbf{e}^i, q)} + \sum_{q \in Q^{c'}} e^{\mathrm{Sim}(\mathbf{e}^i, q)}},
    \end{split}
\end{equation}
where \(Q^c\) contains class-specific prototypes and \(Q^{c'}\) contains prototypes from other classes.

To ensure consistent feature learning across institutions, we introduce a mean prototype consistency regularization:
\begin{equation}\small
\setlength\abovedisplayskip{0pt} \setlength\belowdisplayskip{0pt}
    \begin{split}
    L_{consis} = \sum_{v=1}^d \left( \left( e^{i,v}_\mathrm{enc} - \bar{q}^{c,v}_\mathrm{enc} \right)^2 + \left( e^{i,v}_\mathrm{dec} - \bar{q}^{c,v}_\mathrm{dec} \right)^2 \right).
    \end{split}
\end{equation}

The final loss combines our multi-level prototype loss (\(L_{\mathrm{MP}} = L_{contra} + L_{consis}\)) with the standard Dice loss:
\begin{equation}\small
\setlength\abovedisplayskip{0pt} \setlength\belowdisplayskip{0pt}
    \begin{split}
    L_{\text{total}} = L_{\mathrm{MP}} + L_{dice}.
    \end{split}
\end{equation}

\subsection{Convergence Analysis}
We establish the theoretical foundations of our method under the following standard assumptions.

\noindent \textbf{Assumption 1 (Lipschitz Smoothness of Loss Functions):} The loss function \(\mathcal{F}(\Theta_e) \) for any client \( e \) satisfies \( L_{sm} \)-smoothness, implying its gradient exhibits Lipschitz continuity with a positive constant \( L_{sm} \). Mathematically, the following inequality holds:  

\begingroup
\setlength\abovedisplayskip{0pt}
\setlength\belowdisplayskip{0pt}
\setlength{\jot}{1pt}

\begin{equation}\small
\begin{split}
F(\Theta)
  &= \frac{1}{N}\sum_{(x,y)}
     \Bigl[L_{\mathrm{dice}}(x,y;\Theta)
          + L_{\mathrm{contra}}(x,y;\Theta)\\[-3pt]
  &\qquad\quad
          + L_{\mathrm{consis}}(x,y;\Theta)\Bigr].
\end{split}
\end{equation}

\endgroup

\noindent \textbf{Assumption 2 (Unbiased Stochastic Gradients with Bounded Dispersion):} Each client \( e \) computes stochastic gradients that are unbiased estimators of the global gradient, with their variance constrained by \( \sigma^2 \):
 \begin{equation}
 \setlength\abovedisplayskip{0pt} \setlength\belowdisplayskip{0pt}
 \begin{split}
 \mathbb{E}[g_{e}] = \nabla \mathcal{F}, \mathbb{E}\|g_{e} - \nabla \mathcal{F}\|^{2} \leq \sigma^{2}.
 \end{split}
 \end{equation}

\noindent \textbf{Assumption 3 (Bounded Prototype Norm):} The uploaded encoder/decoder prototypes \( p^{k}_{\text{dec}} \) from each client satisfy \(|p^{k}_{\text{dec}}| \leq G\), where \(G > 0\) is a predefined constant ensuring stable optimization.

\noindent \textbf{Assumption 4 (Lipschitz Continuity of InfoNCE Loss):} For unit-norm anchor vectors, the InfoNCE loss \( L_{\mathrm{contra}} \) with respect to the positive sample \( r \) satisfies the Lipschitz gradient condition:
  \begin{equation}
  \label{tau}
   \setlength\abovedisplayskip{0pt} \setlength\belowdisplayskip{0pt}
   \begin{split}
  \left|\frac{\partial L_{\mathrm{contra}}}{\partial r}\right| \leq \frac{1}{\tau}. 
  \end{split}
  \end{equation}

\noindent \textbf{Theorem 1 (One-Round Convergence Bound).} Under Assumptions 1-4, for any client participating in the Federated Learning process, the expected objective function after each communication round satisfies:
\begingroup
  \setlength\abovedisplayskip{0pt}
  \setlength\belowdisplayskip{0pt}
  \setlength{\jot}{1pt}
  \begin{equation}\label{th1}\small
  \begin{split}
  \mathbb{E}\!\bigl[\mathcal{F}_{t+1}\bigr]
    &\le \mathcal{F}_t
           -\alpha(\eta)\sum_{e=0}^{E-1}
           \|\nabla\mathcal{F}(\Theta_e)\|^2\\[-2pt]
    &\quad+ \frac{L_{sm}\eta^2E}{2}\,\sigma^2
            + \lambda_{c}\frac{E\eta G}{\tau}.
  \end{split}
  \end{equation}
\endgroup

This result establishes an upper bound on the expected objective function after each communication round, with convergence guaranteed under proper choice of $\eta$ and $\lambda_c$:
\begin{equation}
\setlength\abovedisplayskip{0pt} \setlength\belowdisplayskip{0pt}
\begin{split}
\eta_{e'}<\frac{2(\sum_{e=0}^{e'}\|\nabla \mathcal{F}(\Theta_e)\|^2-\frac{\lambda_c EG}{\tau})}{L_{sm}(\sum_{e=0}^{e'}\|\nabla \mathcal{F}(\Theta_e)\|^2+E\sigma^2)},
\end{split}
\end{equation}
where $e^{\prime }= 1/ 2, 1, \ldots , E- 1, $ and
\begin{equation}
\setlength\abovedisplayskip{0pt} \setlength\belowdisplayskip{0pt}
\begin{split}
\lambda_t<\frac{\tau \|\nabla \mathcal{F}(\Theta_e)\|^2}{EG}.
\end{split}
\end{equation}

\noindent \textbf{Theorem 2 (Non-convex convergence rate).} Under Assumptions 1-4, and $\Delta=\mathcal{F}_0-\mathcal{F}^*$ where $\mathcal{F}^*$ refers to the local optimum. For an arbitrary client, given any $\epsilon > 0$, after
\begin{equation}T=\frac{2\Delta}{E\epsilon(2\eta-L_{sm}\eta^2)-E\eta(L_{sm}\eta\sigma^2+\frac{2\lambda_c G}{\tau})}\end{equation}
communication rounds, we have
\begin{equation}
\setlength\abovedisplayskip{0pt} \setlength\belowdisplayskip{0pt}
\begin{split}
\frac{1}{TE}\sum_{t=0}^{T-1}\sum_{e=0}^{E-1}\mathbb{E}[\|\nabla \mathcal{F}(\Theta_e)\|^2]<\epsilon,
\end{split}
\end{equation}
if
\begin{equation}
\setlength\abovedisplayskip{0pt} \setlength\belowdisplayskip{0pt}
\begin{split}
\eta<\frac{2(\epsilon-\frac{\lambda_c G}{\tau})}{L_{sm}(\epsilon+\sigma^2)}\quad and\quad \lambda_c<\frac{\tau\epsilon}{G}.
\end{split}
\end{equation}
The result shows that with proper hyperparameters $(\eta, \lambda_c)$, the expected norm of gradients can be bounded within any given precision level $\epsilon > 0$. \textbf{Please refer to the appendix for detailed evidence and analysis.}

\begin{table*}[t]
\centering
\scriptsize
\resizebox{\linewidth}{!}{
  \setlength\tabcolsep{3.5pt}
  \renewcommand\arraystretch{1.1}
  \begin{tabular}{r||ccccIccIccccccIcc}
    \hline\thickhline
    %%%%% 上半部分：DSC 指标结果 %%%%%
    \rowcolor{mygray}
    & \multicolumn{6}{cI}{Histology nuclei segmentation}
    & \multicolumn{8}{c}{Prostate MRI segmentation}\\
    \cline{2-15}
    \rowcolor{mygray}
    \multirow{-2}{*}{Methods} 
    & TCIA & CRC & KIRC & TNBC & AVG & $\triangle$
    & BIDMC & HK & HCRUDB & RUNMC & BMC & UCL & AVG & $\triangle$ \\
    \hline\hline
    \fedavg{} \pub{ASTAT17} 
    & 72.92 & 73.74 & 73.60 & 57.72 & 69.50 & - 
    & 68.81 & 79.86 & 86.85 & 80.39 & 82.23 & 74.90 & 78.80 & -\\

    \fedprox{} \pub{arXiv20} 
    & 72.53 & 74.00 & 74.07 & 55.60 & 69.00 & -0.50
    & 68.44 & 82.06 & 86.01 & 80.38 & 81.53 & 70.44 & 78.10 & -0.70\\

    \moon{} \pub{CVPR21}
    & 72.66 & 73.81 & 73.53 & 57.25 & 69.30 & -0.20
    & 70.76 & 84.39 & 85.89 & 81.32 & 82.86 & 71.51 & 79.50 & +0.70\\

    \fedproto{} \pub{AAAI22} 
    & 71.66 & 71.79 & 73.31 & 56.22 & 68.20 & -1.30
    & 67.06 & 79.30 & 85.06 & 78.99 & 80.97 & 71.22 & 77.10 & -1.70\\

    \harmof{} \pub{AAAI22} 
    & 76.77 & 76.02 & 74.72 & 60.09 & \underline{71.90} & \underline{+2.40}
    & 79.66 & 80.81 & 87.88 & 78.34 & 83.51 & 76.70 & \underline{81.20} & \underline{+2.40}\\

    \FPL{} \pub{CVPR23}
    & 74.02 & 75.12 & 73.36 & 64.16 & 71.66 & +2.16
    & 67.01 & 83.96 & 83.08 & 80.51 & 80.82 & 69.14 & 77.40 & -1.40\\

    FedContrast{} \pub{MICCAI23} 
    & 75.98 & 73.69 & 72.86 & 59.38 & 70.50 & +1.00
    & 65.43 & 83.87 & 83.73 & 80.20 & 79.74 & 66.29 & 76.50 & -2.30\\

    \FedPLVM{} \pub{NeurIPS24} 
    & 73.46 & 75.58 & 72.23 & 62.60 & 71.00 & +1.50
    & 67.41 & 80.70 & 84.73 & 79.59 & 81.28 & 71.17 & 77.50 & -1.30\\

    \feduv{} \pub{CVPR24} 
    & 72.71 & 74.74 & 73.01 & 59.48 & 70.00 & +0.50
    & 72.40  &82.12 &86.04&78.56 & 82.23 &69.59  & 78.50 & -0.30\\
    \hline
    \rowcolor[HTML]{FFF0C1}
    \textbf{FedBCS}
    & 76.21 & 78.70 & 74.42 & 67.00 & \textbf{74.10} & \textbf{+4.60} 
    & 77.56 & 84.56 & 87.22 & 83.31 & 85.01 & 77.93 & \textbf{82.60} & \textbf{+3.80}\\
  \end{tabular}
}
% \vspace{-10pt}
\captionsetup{font=small}
\caption{\small{
\textbf{Comparison with the SOTA methods} on Histology nuclei segmentation and Prostate MRI segmentation tasks. AVG denotes average accuracy calculated on all domains. Best in bold and second with underline. These notes are the same to others.
}}
\label{tab:combined}
% \vspace{-5pt}
\end{table*}

\begin{table}[t]\small
\centering

% 上半部分表格（保持原样）
\resizebox{\columnwidth}{!}{
\setlength\tabcolsep{4.5pt}
\begin{tabular}{cc||ccccIc}
\hline \thickhline
\rowcolor{mygray}
&& \multicolumn{5}{c}{Histology nuclei segmentation}\\
\cline{3-7}
\rowcolor{mygray}
\multirow{-2}{*}{\cfdlplabbv{}}& \multirow{-2}{*}{\lfbsnabbv{}} 
& TCIA & CRC & KIRC & TNBC & AVG \\
\hline\hline
& & 72.92 & 73.74 & 73.60 & 57.72 & 69.50\\ 
\ding{51} & & 75.77 & 73.92 & 73.74 & 65.38 & 72.20\\ 
& \ding{51} & 74.80 & 76.45 & 74.65 & 63.14 & 72.30 \\
\rowcolor[HTML]{D7F6FF}
\ding{51} & \ding{51} & 76.21 & 78.70 & 74.42 & 67.00 & \textbf{74.10}\\
% \thickhline
\end{tabular}
}

% 下半部分表格（修复背景颜色）
% \vspace{1mm}
\resizebox{\columnwidth}{!}{
\setlength\tabcolsep{3.5pt}
\begin{tabular}{cc||ccccccIc}
\hline \thickhline
\rowcolor{mygray}
&& \multicolumn{7}{c}{Prostate MRI segmentation}\\ % 修正为7列合并
\cline{3-9} % 修正为3-9
\rowcolor{mygray}
\multirow{-2}{*}{\cfdlplabbv{}}& \multirow{-2}{*}{\lfbsnabbv{}} 
& BIDMC & HK & HCRUDB & RUNMC & BMC & UCL & AVG \\
\hline\hline
& & 68.81 & 79.86 & 86.85 & 80.39 & 82.23 & 74.90 & 78.80\\ 
\ding{51} & & 72.73 & 86.46 & 88.54 & 83.94 & 83.78 & 75.82 & 81.90\\ 
& \ding{51} & 74.53 & 83.58 & 86.59 & 79.29 & 82.33 & 74.13 & 80.10\\ 
\rowcolor[HTML]{D7F6FF}
\ding{51} & \ding{51} & 77.56 & 84.56 & 87.22 & 83.31 & 85.01 & 77.93 & \textbf{82.60}\\ 
% \hline \thickhline
\end{tabular}
    }
\caption{\small\textbf{Ablation study of key components} of our method in Histology nuclei segmentation and Prostate MRI segmentation task.  Please see \cref{sec:ablation} for details.
}
\label{tab:ablation}
\end{table}

\begin{table}[t]\small
\centering
\resizebox{\columnwidth}{!}{
\setlength\tabcolsep{3.5pt}
\begin{tabular}{c||cccccc||c} % 移除第一列后的列定义
\hline \thickhline
\rowcolor{mygray}
& \multicolumn{6}{c||}{Avg \# of Prototypes Uploaded} & \\ 
\cline{2-7} % 调整横线范围
\rowcolor{mygray}
\multirow{-2}{*}{Methods} % 仅保留UPCR列
& BIDMC & HK & HCRUDB & RUNMC & BMC & UCL & \multirow{-2}{*}{AVG} \\ 
\thickhline 
FedProto & 2.00 & 2.00  & 2.00  & 2.00  & 2.00  & 2.00  & 77.10\\ 
FPL & 2.00  & 2.00  & 2.00  & 2.00  & 2.00  & 2.00  & 77.40\\ 
FedContrast & 9.00 & 9.00 & 9.00 & 9.00 & 9.00 & 9.00 & 76.50\\ 
FedPLVM & 212.10 & 128.44 & 365.09 & 323.39 & 306.84 & 140.14 & 77.50\\ 
\rowcolor[HTML]{D7F6FF}
\textbf{FedBCS} & 4.00 & 4.00 & 4.00 & 4.00 & 4.00 & 4.00 & \textbf{82.60}\\ 
\thickhline
\end{tabular}
}
\caption{\small \textbf{Comparison of communication costs} with other federated prototype learning methods in prostate MRI segmentation task. \textbf{Avg \# of prototypes uploaded} denotes the average number of prototypes each client sends to the server per training epoch.}
\label{tab:proto}
% \vspace{-10pt}
\end{table}

\section{Experiments}
\subsection{Experiment Settings}
\noindent \textbf{Datasets}. We evaluate methods on two segmentation tasks.
\begin{itemize}[leftmargin=*]
        \setlength{\itemsep}{0pt}
	\setlength{\parsep}{-2pt}
	\setlength{\parskip}{-0pt}
	\setlength{\leftmargin}{-10pt}
	% \vspace{-4pt}
\item \textbf{Histology nuclei segmentation}: We assembled four public datasets (TCIA~\cite{hou2020dataset}, CRC~\cite{awan2017glandular}, KIRC~\cite{irshad2014crowdsourcing}, TNBC~\cite{naylor2018segmentation}) for binary segmentation of cell nuclei in histology images.
\item \textbf{Prostate MRI segmentation}: We used a multi-site dataset from \cite{liu2020ms}, which comprises six sources from three public datasets \cite{lemaitre2015computer,litjens2014evaluation,nciisbi2013} for binary segmentation of the prostate.
% \vspace{-4pt}
\end{itemize}
We treat each data source as a separate client. Due to variations in pathological staining and MRI scanning protocols, images from each client differ visually, yet they maintain a consistent label distribution. The example cases in each domain are presented in Fig.~\ref{fig:problem}. Following \cite{wang2023black,liu2020ms}, we adopt the same preprocessing methods and data splits, and directly evaluate the results on the test sets. For more dataset details, please refer to Appendix.
% For more dataset details, please refer to Appendix xxx.

\noindent \textbf{Model}. For both segmentation tasks, we use commonly used UNet~\cite{ronneberger2015u} as the shared model architecture for training.

% \noindent \textbf{Comparison Methods}. We compare our method against several SOTA federated learning approaches designed to construct a shared global model, including FedAvg (AISTATS’17~\cite{FedAvg_AISTATS17}), FedProx (arXiv’18~\cite{FedProx_MLSys2020}), MOON (CVPR’21~\cite{MOON_CVPR21}), FedProto (AAAI’22~\cite{FedProto_AAAI22}), HarmoFL (AAAI’22~\cite{jiang2022harmofl}), FPL (CVPR’23~\cite{FPL_CVPR23}), FedContrast (MICCAI’23~\cite{FedContrast-GPA_MICCAI23}), FedPLVM (NeurIPS’24~\cite{FedPLVM_NeurIPS24}) and FedUV (CVPR'24~\cite{FedUV_CVPR24}).
\noindent \textbf{Comparison Methods}. We compare our method against several SOTA federated learning approaches designed to construct a shared global model, including FedAvg~\cite{FedAvg_AISTATS17}, FedProx~\cite{FedProx_MLSys2020}, MOON ~\cite{MOON_CVPR21}, FedProto~\cite{FedProto_AAAI22}, HarmoFL~\cite{jiang2022harmofl}, FPL~\cite{FPL_CVPR23}, FedContrast~\cite{FedContrast-GPA_MICCAI23}, FedPLVM~\cite{FedPLVM_NeurIPS24} and FedUV~\cite{FedUV_CVPR24}.

\noindent \textbf{Implementation Details}. All approaches are implemented under identical configurations, using 400 communication rounds and 1 local epoch. For histology nuclei segmentation, we adopt the SGD optimizer with a learning rate of 0.01, while for prostate MRI segmentation we use the Adam optimizer with a learning rate of 1e-4. The optimizer parameters are set with a weight decay of 1e-4, and the training batch size is 6. Detailed hyper-parameter settings are provided in Sec.\ref{sec:ablation}. 

\noindent \textbf{Evaluation Metric}. Following~\cite{wang2023black}, we adopt the widely used Dice score as the evaluation metric for both segmentation tasks. Our evaluation focuses solely on the foreground Dice score. To ensure reproducibility, we fix the random seed for all experiments.

\subsection{Diagnostic Analysis}
\label{sec:ablation}
\begin{figure}[t]
    \begin{minipage}[b]{0.495\linewidth}
        \centering
        \includegraphics[width=\linewidth]{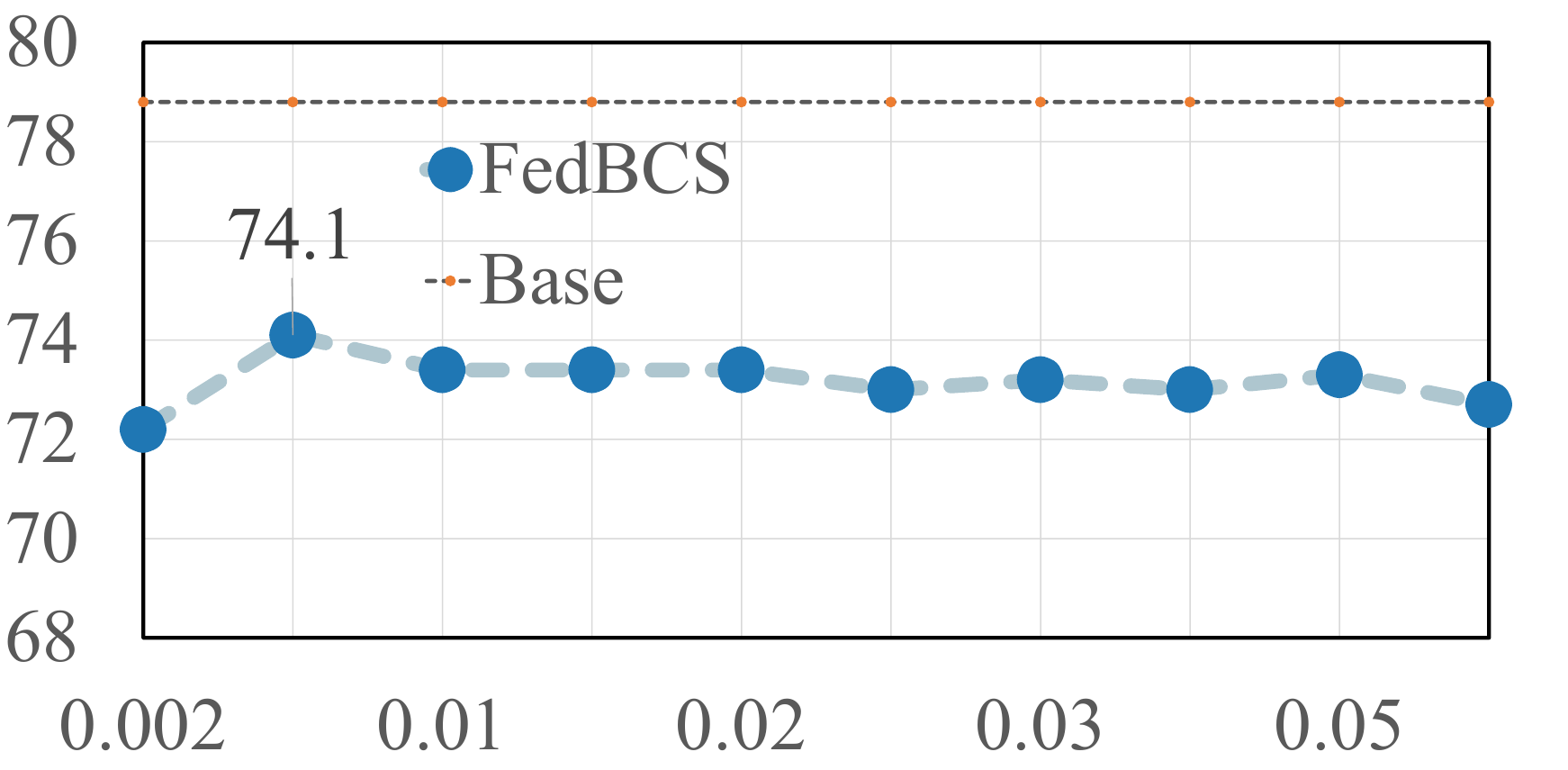}
        \centerline{(a)}
    \end{minipage}
    \begin{minipage}[b]{0.495\linewidth}
        \centering
        \includegraphics[width=\linewidth]{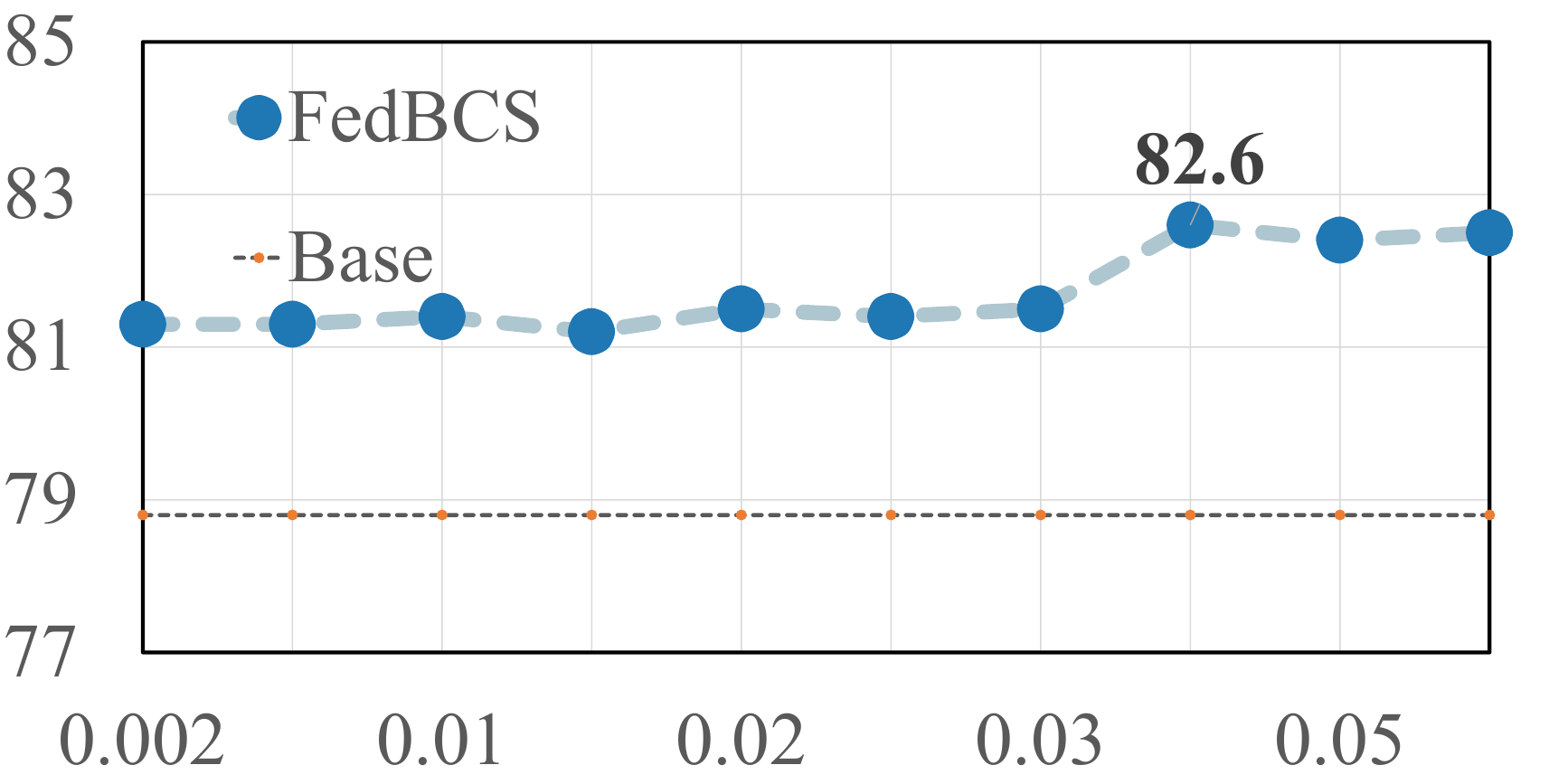}
        \centerline{(b)}
    \end{minipage}
    % \vspace{5pt}
	\captionsetup{font=small}    
	\caption{\small{\textbf{Analysis of FedBCS with different temperature (\cref{tau})}. ``Base" denotes \fedavg{}. See details in \cref{sec:ablation}.
	}}
    % \vspace{-5pt}
	\label{fig:ablation}
\end{figure}

\noindent \textbf{Hyper-parameter Study}. Figure~\ref{fig:ablation} illustrates the impact of hyperparameter $\tau$ (\cref{tau}) on performance. For histology nuclei segmentation, optimal performance is achieved at $\tau$=0.005, while for prostate MRI segmentation the best results occur at $\tau$=0.4. The results, depicted in Fig.~\ref{fig:ablation}, demonstrate that our method consistently surpasses the second-place method across a range of temperatures (refer to Tab.~\ref{tab:combined}).

\begin{figure}[t]
    \centering
    \begin{minipage}[b]{0.493\linewidth}
        \centering
        \includegraphics[width=\linewidth]{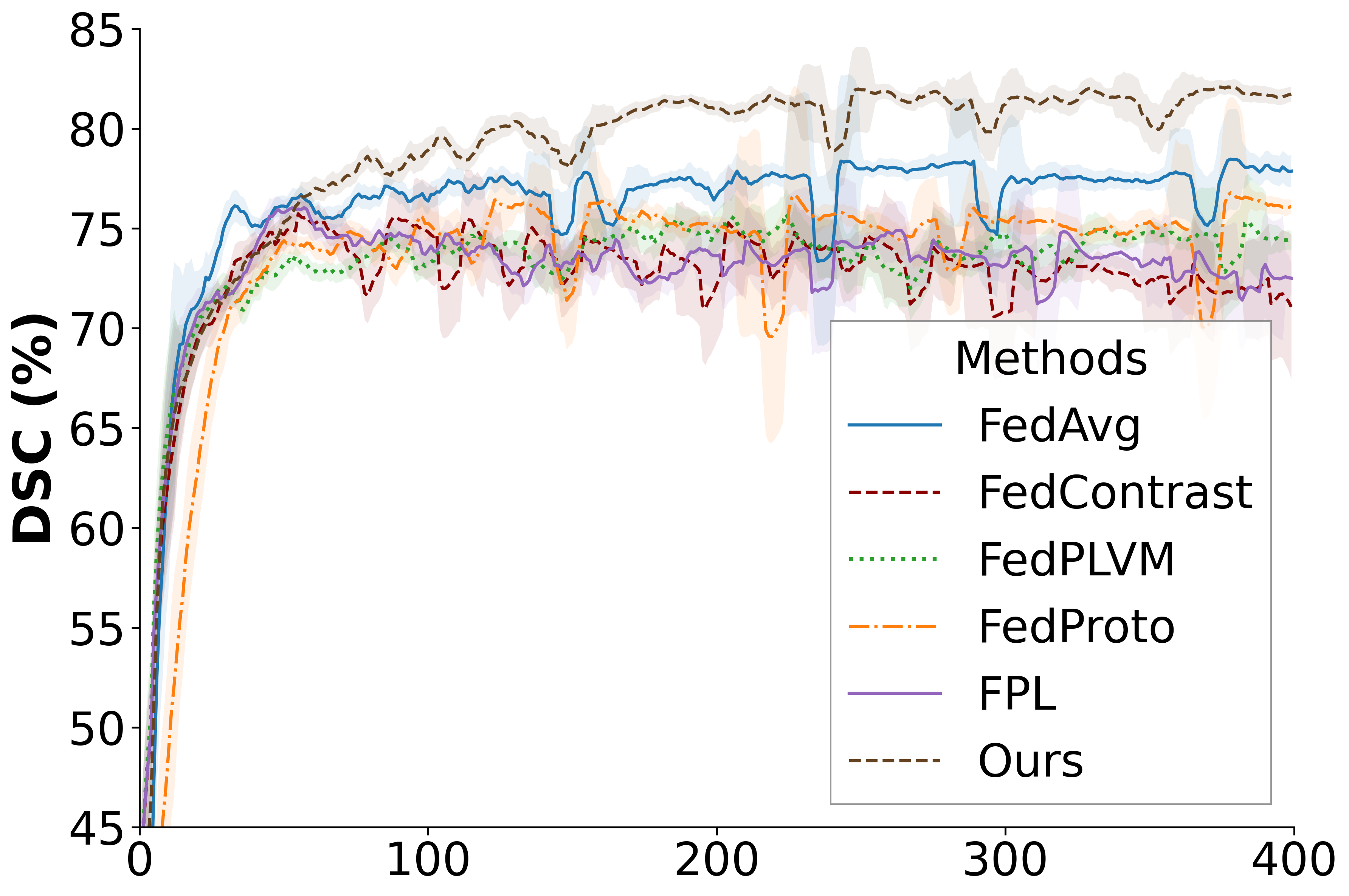}
        \centerline{(a)}
    \end{minipage}
    \begin{minipage}[b]{0.493\linewidth}
        \centering
        \includegraphics[width=\linewidth]{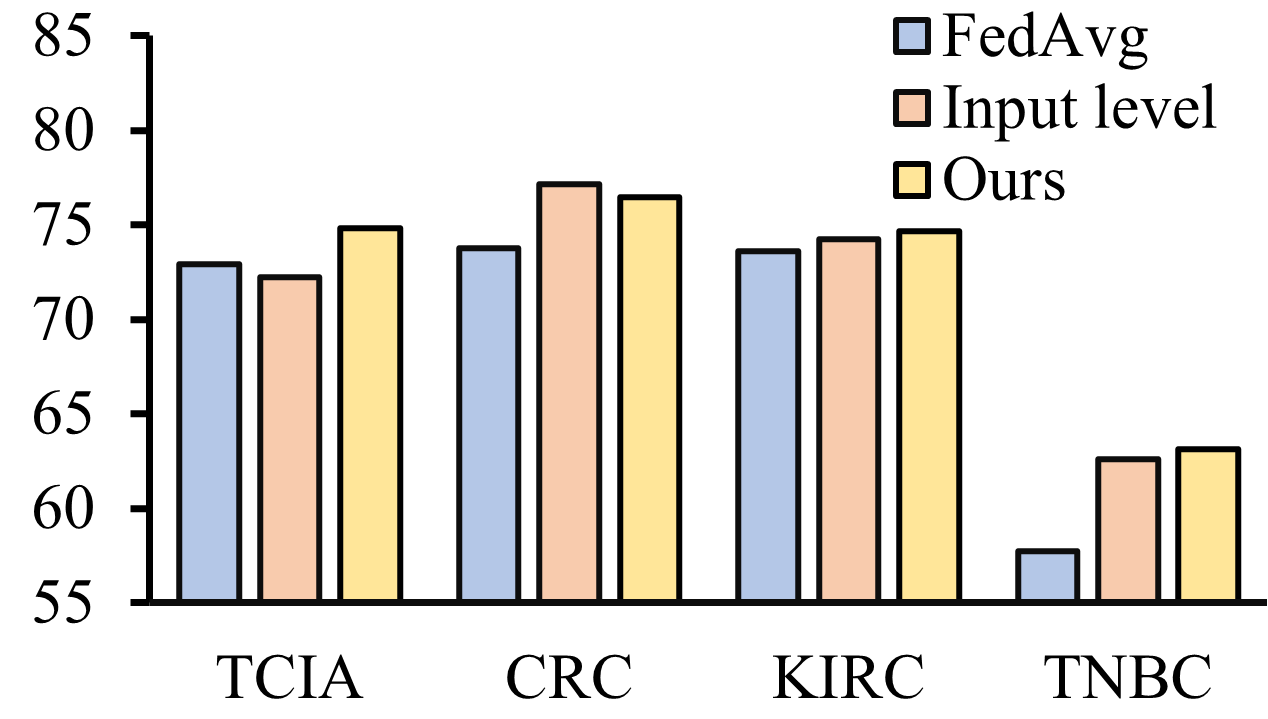}
        \centerline{(b)}
    \end{minipage}
    % \vspace{-5pt}
	\captionsetup{font=small}    
    \caption{\small{\textbf{Performance analysis.} (a) Convergence comparison of differentmethods across communication rounds. (b) Effectiveness of FSR in addressing layerwise style bias compared to input-level normalization. See details in \cref{sec:ablation}.}}
    % \vspace{-10pt}
	\label{fig:amplitude}
\end{figure}

\noindent \textbf{Ablation Study}. To comprehensively analyze the effectiveness of the two components, we conducted an ablation study on both datasets, as shown in Table~\ref{tab:ablation}. The quantitative findings confirm that each component individually enhances performance, and their combined application leads to the optimal outcomes.

\noindent \textbf{Communication Cost Analysis}. We compare our method's communication overhead with other federated prototype approaches in Tab.~\ref{tab:proto} (prostate MRI segmentation) and Fig.~\ref{fig:ablation} (a). As shown in Fig.~\ref{fig:ablation}(a), our method achieves better performance while maintaining stable convergence. In terms of communication efficiency (Tab.~\ref{tab:proto}), FedProto~\cite{FedProto_AAAI22} and FPL~\cite{FPL_CVPR23} generate two prototypes by averaging each class's features, while FedContrast~\cite{FedContrast-GPA_MICCAI23} and FedPLVM~\cite{FedPLVM_NeurIPS24} require more prototypes due to local clustering. Although our method leverages multi-level features to address incomplete contextual representation, we maintain efficiency by uploading only two prototypes per class. The results demonstrate that our approach achieves superior segmentation performance without substantial communication overhead.

\noindent \textbf{Effectiveness of FSR in Addressing Style Bias}. Figure~\ref{fig:amplitude} (b) compares our FSR method with input-level amplitude normalization based on FedAvg, evaluated on the Histology nuclei segmentation dataset with pronounced style differences. Instance normalization was consistently applied in all experiments. The results demonstrate that FSR, by addressing style bias accumulation across layers, achieves better performance than methods that only handle style variations at the input level.

\subsection{Comparison to State-of-the-Arts}
As evidenced by the comparative analysis in Table~\ref{tab:combined}, our framework demonstrates significant improvements over existing approaches, validating FedBCS's enhanced cross-domain generalization capability. Furthermore, the communication efficiency benchmarks in Table~\ref{tab:proto} reveal that FedBCS maintains superior segmentation performance while incurring only marginal communication overhead.

\section{Conclusion}
% In this paper, we addressed two critical challenges in federated medical image segmentation: layerwise style bias accumulation from diverse imaging protocols and incomplete contextual representation learning. Our solution comprises two key components: frequency-domain style recalibration to systematically address style variations across network layers, and context-aware dual-level prototype alignment to capture comprehensive semantic information. Through extensive experiments on two public medical datasets, we demonstrated that this approach effectively improves cross-domain generalization and segmentation accuracy. 
% In this paper, we addressed two critical challenges in federated medical image segmentation: layerwise style bias from diverse protocols and incomplete contextual learning. Our solution comprises frequency-domain style recalibration and context-aware dual-level prototype alignment. Experiments on two datasets demonstrate improved generalization and accuracy.
In this paper, we addressed two critical challenges in federated medical image segmentation: layerwise style bias accumulation and incomplete contextual learning. Our solution comprises frequency-domain style recalibration to address style variations, and context-aware dual-level prototype alignment to capture comprehensive semantics. Experiments on two datasets demonstrate improved performance.

\section*{Acknowledgement}
This work is supported by the National Key Research and Development Program of China (2023YFC2705700), National Natural Science Foundation of China under Grant (62361166629, 62225113, 623B2080), the Major Project of Science and Technology Innovation of Hubei Province (2024BCA003, 2025BEA002), and the Innovative Research Group Project of Hubei Province under Grants 2024AFA017. The supercomputing system at the Supercomputing Center of Wuhan University supported the numerical calculations in this paper.

\bibliography{aaai2026}
\setlength{\leftmargini}{20pt}
\makeatletter\def\@listi{\leftmargin\leftmargini \topsep .5em \parsep .5em \itemsep .5em}
\def\@listii{\leftmargin\leftmarginii \labelwidth\leftmarginii \advance\labelwidth-\labelsep \topsep .4em \parsep .4em \itemsep .4em}
\def\@listiii{\leftmargin\leftmarginiii \labelwidth\leftmarginiii \advance\labelwidth-\labelsep \topsep .4em \parsep .4em \itemsep .4em}\makeatother

\setcounter{secnumdepth}{0}
\renewcommand\thesubsection{\arabic{subsection}}
\renewcommand\labelenumi{\thesubsection.\arabic{enumi}}

\newcounter{checksubsection}
\newcounter{checkitem}[checksubsection]

\newcommand{\checksubsection}[1]{%
  \refstepcounter{checksubsection}%
  \paragraph{\arabic{checksubsection}. #1}%
  \setcounter{checkitem}{0}%
}

\newcommand{\checkitem}{%
  \refstepcounter{checkitem}%
  \item[\arabic{checksubsection}.\arabic{checkitem}.]%
}
\newcommand{\question}[2]{\normalcolor\checkitem #1 #2 \color{blue}}
\newcommand{\ifyespoints}[1]{\makebox[0pt][l]{\hspace{-15pt}\normalcolor #1}}

\end{document}